\documentclass[sn-basic]{sn-jnl}% Basic Springer Nature Reference Style/Chemistry Reference Style
\jyear{2022}%

\theoremstyle{thmstyleone}%
\theoremstyle{thmstyletwo}%

\theoremstyle{thmstylethree}%

\graphicspath{
{figures/Ant/}
{figures/Hopper/}
}

\begin{document}

\title[Invariance to Quantile Selection in Distributional Continuous Control]{Invariance to Quantile Selection in Distributional Continuous Control}


\author*[1,3]{Felix Grün\email{felix.gruen@hs-ruhrwest.de}}
\author[1]{Muhammad Saif-ur-Rehman\email{Muhammad.Saif-ur-Rehmann@hs-ruhrwest.de}}
\author[2]{Tobias Glasmachers\email{tobias.glasmachers@ini.rub.de}}
\author[1]{Ioannis Iossifidis\email{iossifidis@hs-ruhrwest.de}}
\affil*[1]{\orgdiv{Institute of Computer Science}, Ruhr West University of Applied Sciences, Mülheim an der Ruhr, Germany}
\affil[2]{\orgdiv{Institute for Neural Computation}, Ruhr University, Bochum, Germany}
\affil[3]{\orgdiv{Faculty of Electrical Engineering and Information Technology}, Ruhr-University, Bochum, Germany}

% \author*[1,2]{\fnm{First} \sur{Author}}\email{iauthor@gmail.com}



% \affil*[1]{\orgdiv{Department}, \orgname{Organization}, \orgaddress{\street{Street}, \city{City}, \postcode{100190}, \state{State}, \country{Country}}}

% \affil[2]{\orgdiv{Department}, \orgname{Organization}, \orgaddress{\street{Street}, \city{City}, \postcode{10587}, \state{State}, \country{Country}}}

% \affil[3]{\orgdiv{Department}, \orgname{Organization}, \orgaddress{\street{Street}, \city{City}, \postcode{610101}, \state{State}, \country{Country}}}


\abstract{
In recent years distributional reinforcement learning has produced many state of the art results. Increasingly sample efficient Distributional  algorithms for the discrete action domain have been developed over time that vary primarily in the way they parameterize their approximations of value distributions, and how they quantify the differences between those distributions. In this work we transfer three of the most well-known and successful of those algorithms (QR-DQN, IQN and FQF) to the continuous action domain by extending two powerful actor-critic algorithms (TD3 and SAC) with distributional critics.
We investigate whether the relative performance of the methods for the discrete action space translates to the continuous case. To that end we compare them empirically on the pybullet implementations of a set of continuous control tasks. 
Our results indicate qualitative invariance regarding the number and placement of distributional atoms in the deterministic, continuous action setting.

}




\maketitle

\section{Introduction}\label{sec:Introduction}

Distributional Reinforcement Learning (RL) methods aim to make better use of the available interactions of the agent with the environment.
They do this by learning the distribution of the amount of reward expected in the future where non-distributional agents usually learn only the expectation of that distribution. 
In recent years, several studies have empirically proven benefits of distributional RL over traditional, also called expected, RL. 
Although the reasons are not yet fully understood, reported learning curves are often steeper and final performance superior to their non-distributional counterparts. 
In the past machine learning research has repeatedly taken inspiration from neuroscience \citep{hassabis_neuroscience-inspired_2017}. 
Yet, distributional RL is an instance of an advancement in machine learning, inspiring the successful search for evidence of a neurological equivalent.

\cite{bellemare_distributional_2017} first popularized the idea that there is inherent value in learning the entire value distribution beyond risk management.
Choosing the best action according to a value function requires a maximization that is simple for a discrete set of actions but quickly becomes prohibitively expensive for continuous actions.
Agents that do not derive their policy directly from a learned value function (or value distribution) directly parameterize the policy and improve it by adjusting the parameters, following the gradient of a performance measure with respect to the policy parameters. These methods are called policy gradient methods. The policy gradient approach can be combined with learning a value function that can then be used as the performance measure, forming an Actor-Critic scheme. 
Value based methods have been drastically improved in the past by using various incarnations of distributional RL. Several significant distributional methods built on DeepMinds Deep Q-Networks \citep[DQN;][]{mnih_human-level_2015}, each surpassing the performance of the previous one. Subsequently, distributional methods were developed that could be applied to continuous action spaces, some of which take inspiration from those DQN based methods. In this work we bring three of those DQN based approaches to the continuous action domain in a way that lets us compare them and see whether their relative performance from the discrete domain transfers. The three algorithms are called Quantile Regression DQN \citep[QR-DQN;][]{dabney_distributional_2018}, Implicit Quantile Networks \citep[IQN;][]{dabney_implicit_2018} and Fully parameterized Quantile Function \citep[FQF;][]{yang_fully_2019}. They have in common that they represent the value distribution by approximating the inverse cumulative distribution function, also called quantile function, with a discrete set of atoms. All three also use quantile regression and train by minimizing a quantile huber loss \citep{kotz_robust_1964}. Although there are other promising approaches to distributional RL \citep[see for example][]{nam_gmac_2021, nguyen-tang_distributional_2021} our choice enables us to investigate the influence of two factors:
The strategy for quantile fraction selection, determining at which points the quantile function should be predicted and
the amount of atoms used, i.e. the resolution of the approximation.

In the following we will first cover the mathematical foundation for distributional RL, then discuss the relation of previously published work to ours in section \ref{sec:Background}. The distributional critics are described in section \ref{sec:Methods}. Section \ref{sec:ImplementationAndExperiments} details the conducted experiments. We present results in section \ref{sec:Results}. Finally, we discuss our results and provide an outlook to possible future work in section \ref{sec:Discussion}.

\section{Background}\label{sec:Background}
We formalize the RL setting as a Markov Decision Process $(S, A, R, P, \gamma)$ where $S$ and $A$ are the state and action spaces, respectively, $R: S\times A \rightarrow R$ is a possibly stochastic reward function that can in general depend on state and action.
$P(  \cdot \mid s, a)$ encodes the environment's transition dynamics.
$\gamma \in [0,1]$ is the discount factor, dictating how farsighted the agent should be.
In our case the agent takes continuous actions $a \in A$, given the current observation of the state of the environment $s \in S$.
The choice of action is governed by a policy $\pi$ with $\sum_a \pi (a\mid s) = 1$.
Overloading notation, a deterministic policy directly maps states to actions: $\pi(s) = a$. The value of a state $s$ or state-action pair $(s, a)$ is defined as the expected (discounted) return starting from $s$ and if specified taking action $a$.
\begin{equation}\label{eq:return}
    v_\pi(s)=\mathbb{E}\left[\sum\limits_{t=t_s}^{T} \gamma^iR(s_{t}, a_{t})\right] %\text{ and } q(s, a)=\sum\limits_{i=0}^{T-t} \gamma^ir_{t+i}$$
\end{equation}
where $s_{t_s}=s$, $a_t\sim\pi (\cdot \mid s_t)$, $s_{t-1}\sim P(\cdot \mid s_t, a_t)$ and $T$\footnote{In this work we only consider episodic environments but in general $T=\infty$ is allowed} is the time step terminating the episode. The bellman equations relate the value of a state to those of following states:
\begin{equation*}
    v_\pi(s) = \sum_{a \in \mathcal{A}_s} \pi(a\mid s) \sum_{s' \in \mathcal{S}} P(s'\mid s, a) \left(R(s, a) + \gamma v_\pi(s')\right) \forall s \in \mathcal{S}
\end{equation*}
%\begin{equation}
%    v_*(s) = \max\limits_{a \in \mathcal{A}_s} \sum_{s' \in \mathcal{S}} P(s'\mid s, a) \left(R(s, a) + \gamma v_*(s')\right) \forall s \in \mathcal{S}
%\end{equation}
\begin{equation}
    \begin{split}
        q_\pi(s, a) = \sum_{s' \in \mathcal{S}} P(s'\mid s, a) & \biggl(R(s, a) + \gamma \sum_{a' \in \mathcal{A}_{s'}} \pi(a'\mid s') q(s', a')\biggr) \\ 
        & \forall s \in \mathcal{S}, a \in \mathcal{A}
    \end{split}
\end{equation}
%\begin{equation}
%    q_*(s, a) = \sum_{s' \in \mathcal{S}} P(s'\mid s, a) \left(R(s, a) + \gamma \max\limits_{a' \in \mathcal{A}_{s'}} q(s', a')\right) \forall s \in \mathcal{S}, a \in \mathcal{A}
%\end{equation}
Applying a Bellman operator $\mathcal{T}$ assigns the right-hand side of these equations to the operand. Both Dynamic Programming and Temporal Difference methods can be straightforwardly derived from these, that learn these value functions analytically or empirically, respectively.

\subsection{Distributional RL}
In contrast, distributional RL algorithms learn the distribution of the random variable that represents the return, which we call $Z$ following the existing literature. $Z(s)$ is equal to the right-hand side of equation \ref{eq:return} without the expectation. Hence, $v(s)=\mathbb{E}\left[Z(s)\right]$. The distributional Bellman operator can be defined as follows:
\begin{equation}
    \mathcal{T}_\pi Z(s, a) = Z(s, a) + \gamma \sum_{s' \in \mathcal{S}} P(s'\mid s, a)\sum_{a' \in \mathcal{A}_{s'}} \pi(a'\mid s') Z(s', a')
\end{equation}
The $p$-Wasserstein distance between two probability distributions $U, V$ is essential for many theoretical results around distributional RL and can be defined as follows:
\begin{equation*}
    W_p(U,V)=\left( \int_{\omega = 0}^1 \lvert F_V^{-1}(\omega) - F_U^{-1}(\omega) \rvert ^p d\omega \right) ^{1/p}
\end{equation*}
Although the distributional bellman operator is a contraction mapping in the maximal form of the Wasserstein metric (\ref{eq:maxWasserstein}) its fixed point cannot be directly approximated via stochastic gradient descent using the Wasserstein metric as a loss function \citep{bellemare_distributional_2017}.
\begin{equation}\label{eq:maxWasserstein}
    \bar{d}_p(Z_1, Z_2):= \sup_{s, a}W_p(Z_1(s, a), Z_2(s, a))
\end{equation}
\citeauthor{dabney_distributional_2018} found a way to circumvent this using quantile regression \citep{koenker_regression_1978}. They prove that the distributional Bellman operator combined with their quantile projection is still a contraction in $\bar{d}_\infty$. This combined operation is practically approximated by minimizing a huber quantile loss:
\begin{equation}
    \rho_\tau^\kappa(u)=\lvert \tau - \delta_{\{u<0\}}\rvert \frac{\mathcal{L}_\kappa (u)}{\kappa}
    \text{ where } \mathcal{L}_\kappa (u) =
    \begin{cases}
        \frac{1}{2}u^2 & \text{if } \lvert u \rvert \leq \kappa \\
        \kappa (\lvert u \rvert - \frac{1}{2}\kappa) & \text{otherwise}
    \end{cases}
\end{equation}
This loss is used in QR-DQN, IQN and FQF, as well as all our derived distributional critics.

\subsection{Related Work}

In this work we integrate distributional RL methods into actor-critic algorithms. That opens two corresponding lines of prior work. A third line of work is concerned with investigating both the benefits of distributional RL as well as how they arise.

\subsubsection{Distributional RL}
To the best of our knowledge the earliest mention of using more than the mean of the return distribution comes from \cite{jaquette_markov_1973}. With the goals of uncertainty estimation or risk-sensitivity in mind, researchers \cite[e.g.][respectively]{engel_reinforcement_2005, morimura_parametric_2012} started using not only the expected value of the return but also the variance. The latter introduces a distributional Bellman equation for conditional return densities. 
The modern line of publications in which distributional RL is used primarily to improve overall performance and not to solve a subproblem, starts with \cite{bellemare_distributional_2017}.
They built on the success of DQN in the Arcade Learning Environment (ALE) to introduce their C51 algorithm (C for categorical, 51 for the number of atoms they ended up using). It uses a fixed support over the range of possible returns, the weights of which are then learned to represent the return density. However, this approach requires either knowing the range of returns a priori or rescaling rewards to fit a predefined range. Another drawback is that their algorithm left a theory-practice gap: It uses the cross-entropy term of the KL divergence as a loss, whereas the motivation was based on the Wasserstein metric.
\cite{dabney_distributional_2018} solved both problems with their Quantile-Regression DQN (QR-DQN). Instead of learning an approximation of the return density function, as C51 does, QR-DQN approximates the inverse cumulative distribution function, also called quantile function. Both algorithms use a fixed number of (discrete) ‘atoms’ for their respective approximation. For QR-DQN, these atoms are distributed equally over the probability space and their location on the return axis can be learned. \cite{dabney_implicit_2018} introduce another algorithm called Implicit Quantile Networks (IQN)  in which the quantiles are not predefined and not equidistant but sampled from some distribution over $[0,1]$. The eponymous network receives as input the current state information and an embedding of the sampled quantiles and outputs the corresponding quantile values. As a result, this approach allows learning the entire quantile function implicitly instead of fixed quantiles. It also accommodates risk-affine or risk-averse policies by using distortion risk measures or sampling the quantiles $\tau$ from non-uniform distributions.
Subsequently, \cite{yang_fully_2019} built on IQN by improving on the purely random choice of quantiles. They combined aspects of C51 and QR-DQN by creating separate artificial neural networks to learn both the quantiles and their corresponding values, in an algorithm called Fully Parameterized Quantile Function (FQF). One of those networks fulfills the same purpose as the IQN but instead of using sampled quantiles, a separate network is trained to select those quantiles that minimize the approximation error that inevitably arises when trying to represent a continuous distribution with a finite number of discrete atoms.
Each of the above mentioned methods reported state-of-the-art performance on the suite of atari games in the ALE at their respective time of publishing, excepting in some cases the Rainbow \cite{hessel_rainbow_2017} algorithm which combined many of the individual improvements to DQN, including the C51 algorithm.

\subsubsection{Continuous Control}
The second line of prior work concerns the handling of continuous action spaces. Although the terms had not been used at the time, policy gradient and actor-critic methods have been introduced as early as the 1970s \cite{witten_adaptive_1977}. Deep RL reached actor-critic methods for the continuous action domain with Deep Deterministic Policy Gradient (DDPG)\cite{lillicrap_continuous_2019}, building on the deterministic policy gradient theorem \cite{silver_deterministic_2014} and its derived algorithms. The best non-distributional, model free algorithms in the continuous action domain are TD3 \cite{fujimoto_addressing_2018} and SAC \cite{haarnoja_soft_2019}. TD3 is based on DDPG and introduces additional critic networks and delayed policy updates to combat overestimation bias. SAC uses maximum entropy RL in an off-policy actor-critic algorithm and, in its current version, a learnable temperature parameter that regulates the influence of the entropy term.

\subsubsection{Distributional Continuous Control}
Several publications built on DDPG \cite[e.g.]{fujimoto_addressing_2018} including \cite{barth-maron_distributed_2018} which introduced Distributed Distributional DDPG (D4PG). In addition to enabling distributed learning, D4PG, like our algorithm, uses distributional RL in an actor-critic scheme but uses the categorical approach used in C51 to learn and represent value distributions. Distributional critics have also been very successfully used in the state-of-the-art algorithm Truncated Quantile Critics (TQC) \cite{kuznetsov_controlling_2020}. The authors use quantile regression critics like ours, pool predicted quantiles from multiple critics and remove the topmost of them to cancel out overestimation bias. As we have done here, they tried both SAC and TD3 as the base algorithm and found SAC to perform better on average. \cite{duan_distributional_2021} present a distributional variant of SAC using sampled quantiles and also report results for fixed and learned quantiles to justify their choice, as well as a TD3 based variant they call TD4. According to their comparison sampled and learned fractions lead to slightly better performance than fixed. We use twice as many random seeds to arrive at a different conclusion and additionally vary the number of atoms used.
\cite{li_distributional_2018} present an algorithm they call DA2C or QR-A2C, a distributional advantage actor-critic that uses Quantile Regression. Even though, reported results in the paper are all in the discrete action domain, it should be straightforward to extend their algorithm to continuous actions.

\subsubsection{Benefits and Analysis of Distributional RL}
Given the impressive performance of modern distributional RL methods it is surprising that the reasons for their empirical superiority are not yet completely clear. Researchers in the RL community have since worked towards understanding the reasons behind the success of distributional RL algorithms. In all of the cases mentioned here the policies derived from a value distribution function or critic only use the mean of the learned distribution. Hence, one way to approach these questions is to determine how and when expected and distributional RL differ. \citeauthor{lyle_comparative_2019} focus on when the approaches differ and rule out several possible reasons for differences through their experimental design. They conclude that in tabular and even many linear value function approximation settings there is no difference in behavior and that differences arise in the presence of nonlinear function approximation. They point out that they look for different behavior, not necessarily superior.
\citeauthor{sun_interpreting_2022} examine the robustness of distributional methods to noisy and adversarial state observations and attest both better theoretical convergence properties, and empirical robustness to distributional algorithms when subjected to noise state information during training.

\section{Methods}\label{sec:Methods}
For this work we implemented three distributional critics:
\begin{itemize}
    \item a Quantile Regression Critic based on \cite{dabney_distributional_2018}
    \item an Implicit Quantile Function Critic based on \cite{dabney_implicit_2018}
    \item a Fully Parameterized Quantile Function Critic based on \cite{yang_fully_2019}.
\end{itemize}

\subsection{Basic Quantile Regression Critic}
After \cite{bellemare_distributional_2017} showed that their distributional Bellman operator is a contraction in a minimal form of the p-Wasserstein metric, but ultimately did not use that property in the C51 algorithm, \cite{dabney_distributional_2018} developed a new method that uses this strong theoretical foundation.
They prove that the combination of the distributional bellman update, and a quantile projection operator is a contraction mapping.
\cite{bellemare_distributional_2017} proved that the Wasserstein distance cannot generally be minimized by stochastic gradient descent(SGD) so \cite{dabney_distributional_2018} make use of a method from economics called Quantile Regression \citep{koenker_regression_1978}. The huber quantile regression loss is asymmetric and convex and allows for unbiased minimization through SGD.
% TODO: something about loss, computation of Q, ... Lτ QR(θ):=E ˆ Z∼Z [ρτ ( ˆ Z − θ)] , where ρτ (u)=u(τ − δ{u<0}), ∀u ∈ R.
Based on Q-learning, the resulting QR-DQN algorithm learns a function $Q: \mathcal{S}\rightarrow Z$ for each action $a\in \mathcal{A}$ that can be used to directly derive a policy by choosing actions $\epsilon$-greedily.
Our distributional critic learns a function $\mathcal{S}\times\mathcal{A}\rightarrow Z$ where $Z$ is the estimated value distribution of taking action $a \in \mathcal{A}$ in state $s \in \mathcal{S}$, represented as a set of quantile values for equidistant fractions.

\subsection{Implicit Quantile Function Critic}
In IQN the quantiles that the learned function predicts are not fixed as in QR-DQN but sampled. As a consequence, the sampled quantiles are additional inputs to the learned function. In addition to improved empirical performance over QR-DQN, this offers the ability to learn risk averse or risk affine policies by sampling the quantiles from non-uniform distributions. 
In the actor-critic framework the learned value distribution is used to update the policy, so although we don’t make use of it in this work this property transfers to the continuous domain.
The quantiles are fed into an embedding network that uses 64 cosine functions. Its outputs are then combined with the state- and, in our work, action information through a Hadamard product.

\subsection{Fully Parameterized Quantile Function Critic}
Representing a generally continuous function (the quantile function, or inverse cumulative distribution function) with a finite set of discrete atoms inevitably results in an approximation error.
What motivates the FQF algorithm is the fact that the extent of this error depends on the choice (i.e. placement) of quantiles that are being predicted. The authors introduce an additional neural network called fraction proposal network (we will write FPN for short). The FPN is being trained in tandem with the prediction network, to select a set of quantile fractions that minimize the approximation error. The proposed fractions are integrated into the state-action features the same way they are in IQN.
While the other two methods compute Q values simply as the mean of the quantile values, FQF uses the following formula:
$$\sum_{i=0}^{N-1}(\tau_{i+1} - \tau_i)F_Z^{-1}\left( \frac{\tau_i + \tau_{i+1}}{2} \right) \text{ with } \tau_0 = 0, \tau_N = 1$$
As the authors point out, for their method the simple mean would be correct only in expectation.

\section{Implementation and Experiments}\label{sec:ImplementationAndExperiments}
\begin{figure}
    \centering
    \includegraphics[width=0.3\textwidth, trim=10cm 7cm 10cm 7cm, clip]{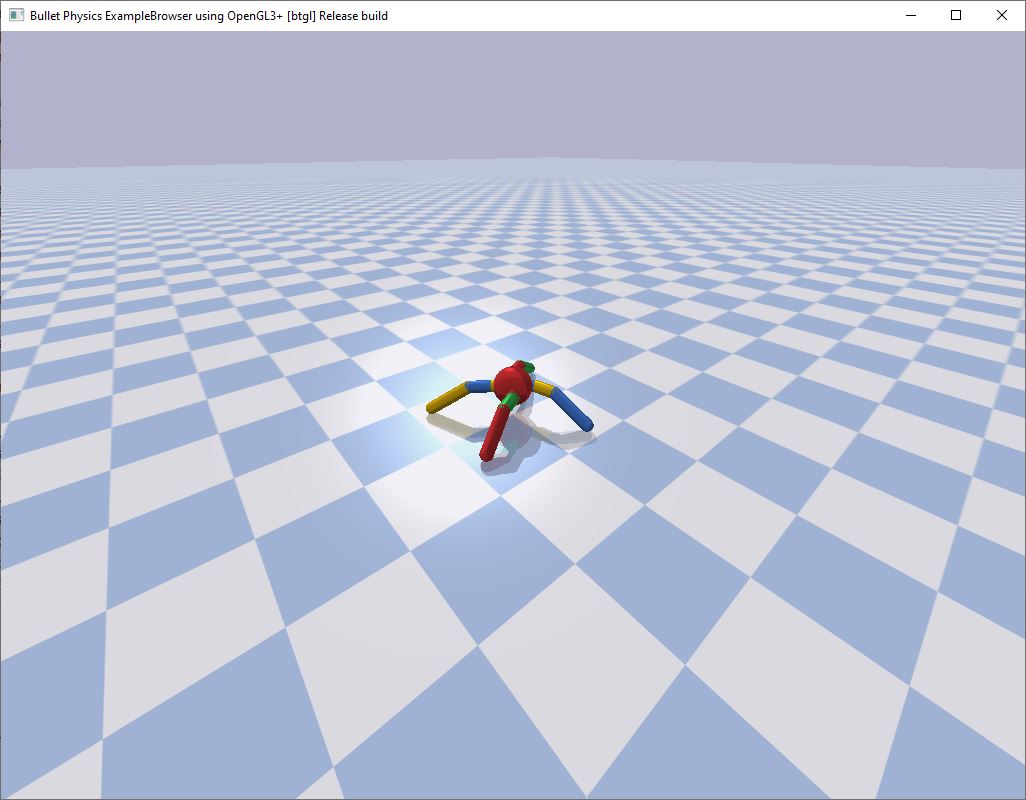}
    \includegraphics[width=0.3\textwidth, trim=10cm 8cm 10cm 6cm, clip]{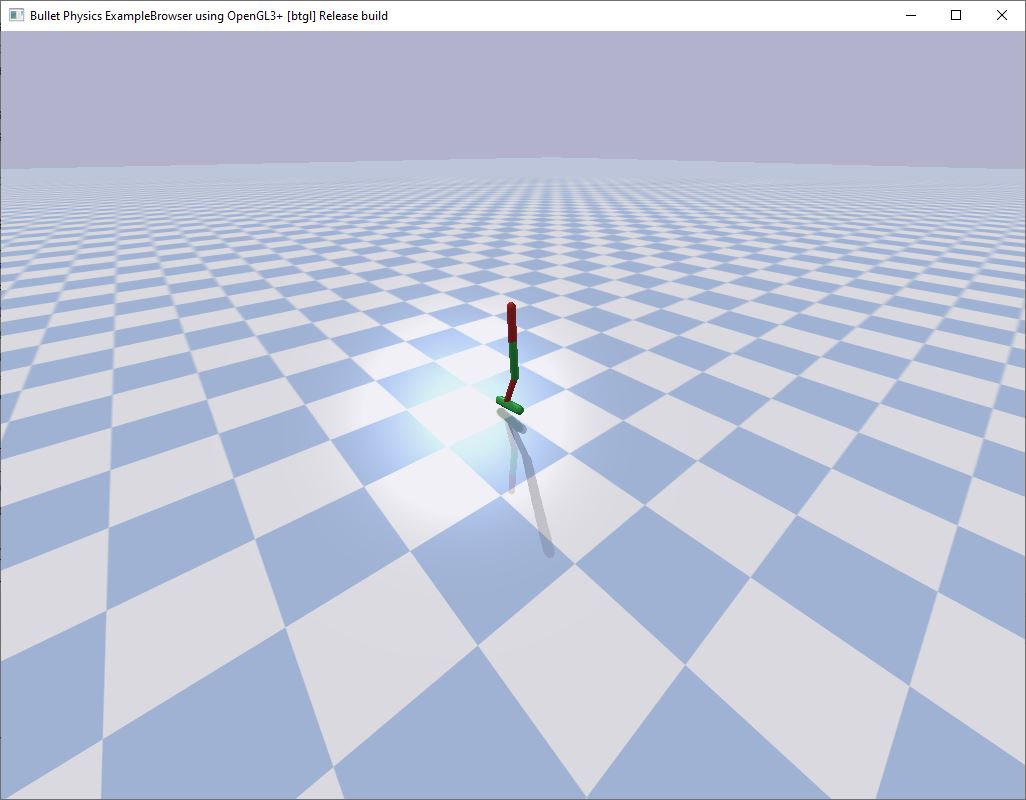}
    \includegraphics[width=0.3\textwidth, trim=10cm 8.5cm 10cm 5.5cm, clip]{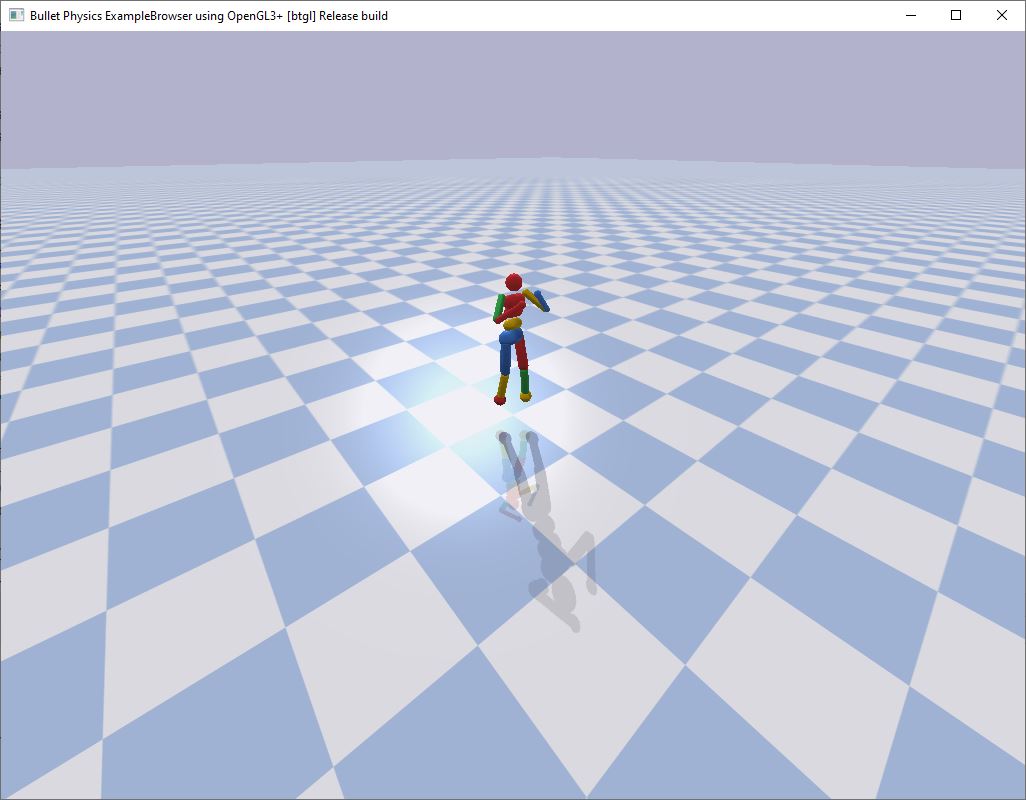}
    \caption{The three benchmark environments used: AntBulletEnv-v0 (left), HopperBulletEnv-v0 (center) and HumanoidBulletEnv-v0 (right)}
    \label{fig:envs}
\end{figure}

We based our implementation on the Stable Baselines3 (SB3)\citep{raffin_stable-baselines3_2021} library. This was done to ensure comparability with the non-distributional versions of TD3 and SAC while using a well-known and stable codebase.
We implemented three distributional critics that could also be used by other algorithms with minor modifications.
The distributional versions of TD3 and SAC using each distributional critic are evaluated and compared with each other as well as the standard, non-distributional versions. We further investigated the effect of varying the “resolution” of the distribution by using different numbers of atoms. 
For all approaches the approximation error should decrease with an increasing number of predicted fractions, at least with equidistant and with uniformly sampled fractions. As the number of quantile fractions goes to infinity the advantage of the learned fractions should vanish. \cite{yang_fully_2019} compare the performance of IQN and FQF using $8, 32 \text{ and } 64$ atoms, showing improved learning speed and performance even for higher resolutions, with learned percentages.
However, at $64$ atoms the difference in performance (mean human normalized scores) is decreased to $5.1\%$ compared to $23\%$ and $33.1\%$ using $8$ and $32$ atoms, respectively. In order to test whether the same is true for continuous actions we conducted all experiments using three \emph{resolutions}.
Given the intuitively and empirically decreasing advantage of FQF with increasing resolution we broadened the range and chose $7$, $51$ and $100$ atoms to see whether the tendency from these three data points holds.
$7$ was chosen as the minimum (over even fewer atoms) to make sure that the distributional aspect of the algorithms takes effect. $100$ was chosen because we deemed it the upper limit of what was computationally feasible for us. Also, it (nearly) leads to a very human set of integer quantile fractions using the basic QR critic. $51$ is chosen because it is the typical choice in distributional RL, used originally in \cite{bellemare_distributional_2017} and adopted for comparability by others \citep[e.g.][]{dabney_distributional_2018}.
For this study, we evaluate and compare our algorithms on a set of pybullet \citep{coumans_pybullet_2016} continuous control tasks. They are the three most commonly used benchmark environments for the evaluation of continuous action agents: Ant, Hopper and Humanoid (see fig. \ref{fig:envs})
This is a subset of the de-facto standard for evaluating continuous control RL algorithms that was first popularized in the MuJoCo \citep{todorov_mujoco_2012} implementation.

For the experiments agents were trained on each environment for $10^6$ steps using the same set of 10 random seeds. 5 evaluation episodes were run after every 1000 steps. We report the means over the 5 evaluation episodes as well as the 10 random seeds, and the standard deviation over the 10 random seeds.

\subsection{Hyperparameter Tuning}
Hyperparameters were tuned solely on the Humanoid environment using the optuna framework \citep{akiba_optuna_2019}. The primary objective behind that choice was finding and using a single set of hyperparameters for all environments. Because the Humanoid environment is the most complex out of the five we hypothesized that it would require the smallest learning rate to be successfully learned. That smaller learning rate could slow down learning in other environments but should allow learning all tasks. Hyperparameters were tuned for each combination of base algorithm, quantile selection strategy and number of quantiles. All variants were given the same budget of runs, i.e. number of tested hyperparameter sets $(105)$. Each hyperparameter set was tried for up to $5 \cdot 10^5$ training steps, subject to optuna’s median pruner after a third of those steps. We focus on a fair comparison rather than reaching the highest possible performance on either algorithm, hence we treat some hyperparameters like the network structure as given and use defaults. In order to achieve a fair and transparent comparison we did not use the best tuning results directly but instead built a single hyperparameter set, adapting only learning rates to the specific configuration. This way it is unlikely that any differences in performance stem from more or less optimal choices of hyperparameter sets. The full list of hyperparameters can be found in appendix \ref{sec:hyperparameters}.

\section{Results}\label{sec:Results}
\iffalse
\begin{figure}
    \centering
    \includegraphics[width=0.45\textwidth]{figures/Ant/100_fqf.pdf}
        \includegraphics[width=0.45\textwidth]{figures/Ant/100_iqn.pdf}\\
    \includegraphics[width=0.45\textwidth]{figures/Ant/100_qr.pdf}
    \includegraphics[width=0.45\textwidth]{figures/Ant/51_fqf.pdf}    
    \caption{Just a Test: Fontsize is OK}
    \label{fig:my_label}
\end{figure}

\begin{figure}
    \centering
    \includegraphics[width=0.3\textwidth]{figures/Ant/100_fqf.pdf}
        \includegraphics[width=0.3\textwidth]{figures/Ant/100_iqn.pdf}
    \includegraphics[width=0.3\textwidth]{figures/Ant/100_qr.pdf}\\
    \includegraphics[width=0.3\textwidth]{figures/Ant/51_fqf.pdf}    
    \includegraphics[width=0.3\textwidth]{figures/Ant/51_iqn.pdf}
    \includegraphics[width=0.3\textwidth]{figures/Ant/51_qr.pdf}    
    \caption{Just a Test: for a 6 plot group the fontsize need to be increased }
    \label{fig:my_label}
\end{figure}

\fi

\begin{figure}
    \centering
    \includegraphics[width=\textwidth]{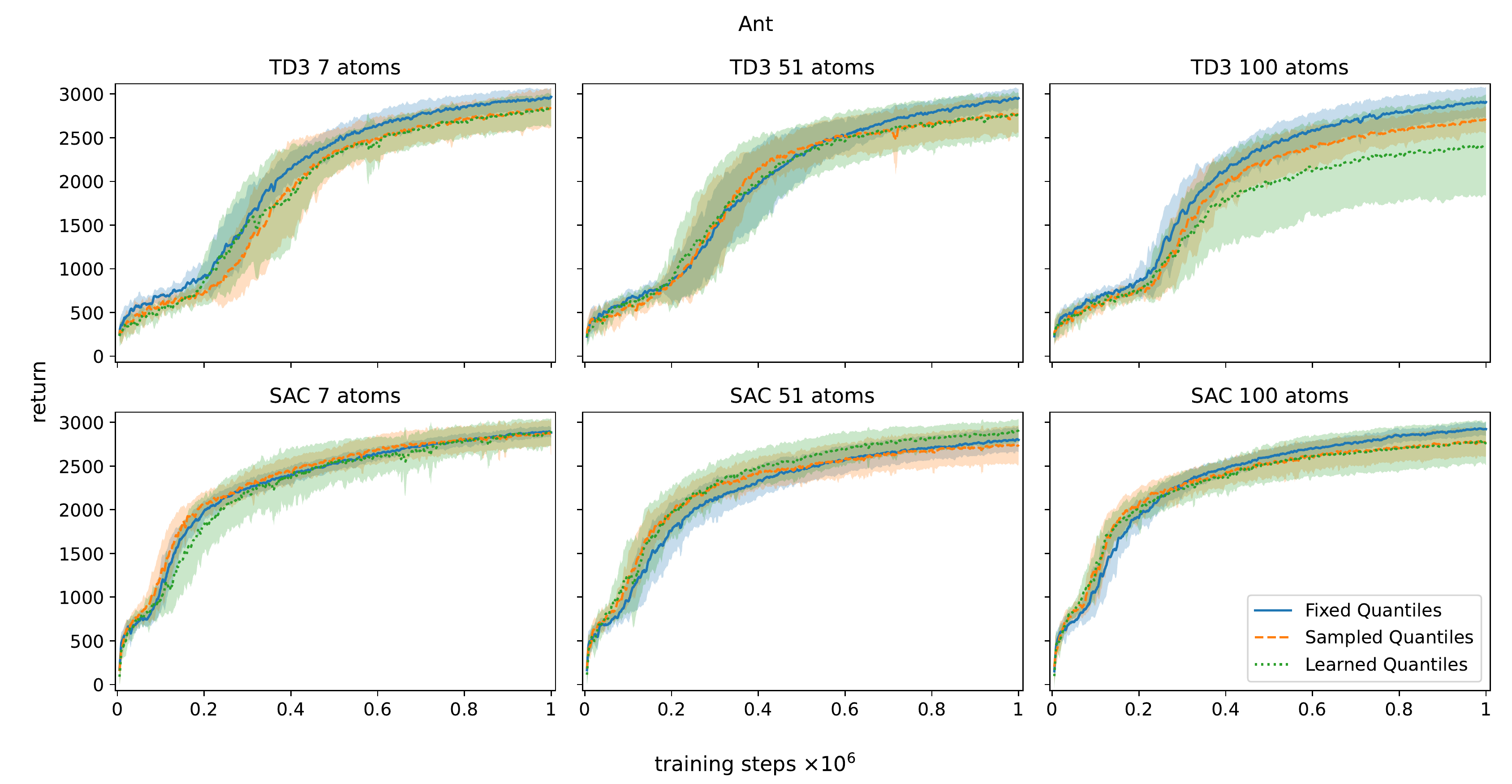}
    \caption{Comparison of the three different quantile fraction selection strategies: fixed (QR-DQN based), sampled (IQN based) and learned (FQF based) in the AntBulletEnv-v0 environment}
    \label{fig:paramCompAnt}
\end{figure}

\begin{figure}
    \centering
    \includegraphics[width=\textwidth]{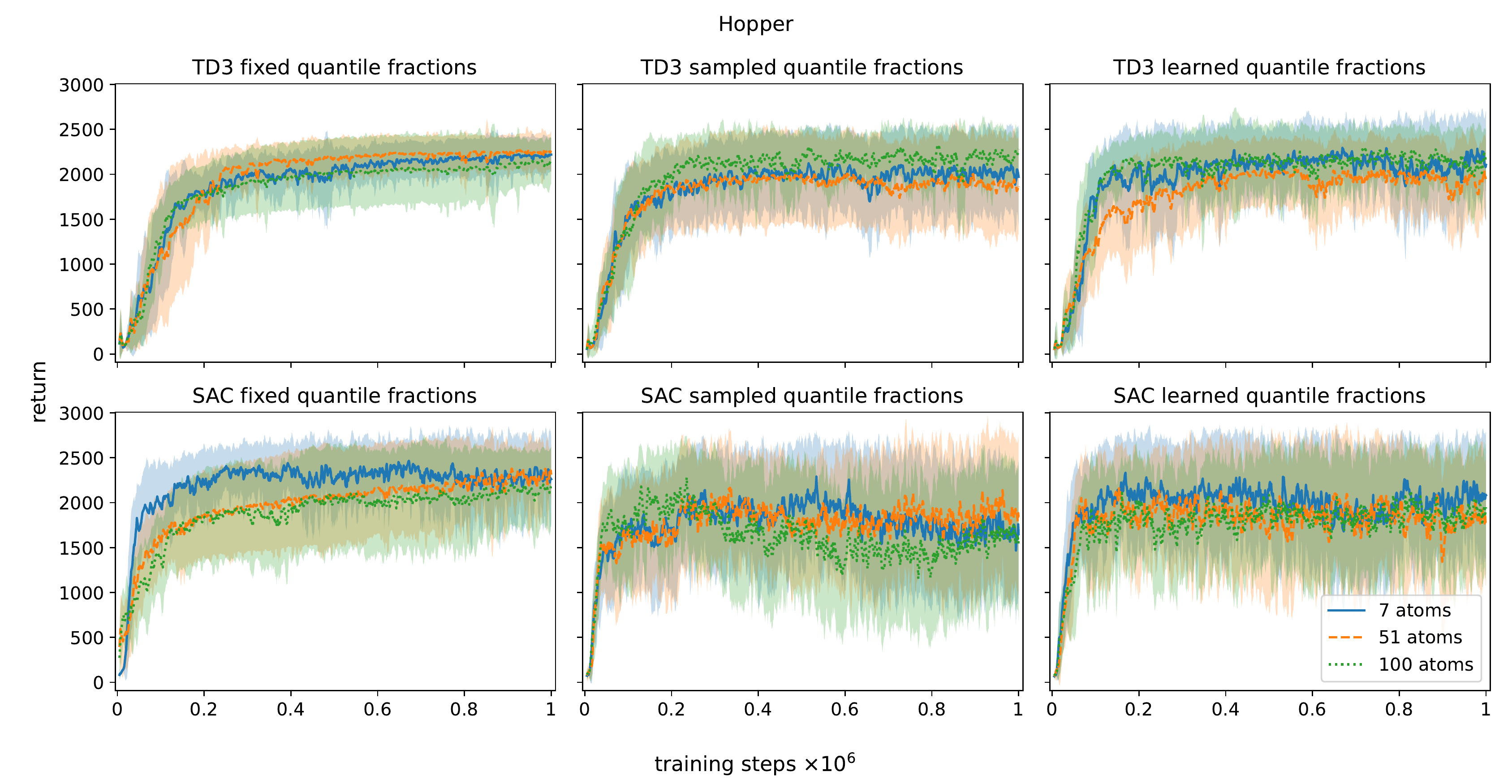}
    \caption{Comparison of different \textit{resolutions} of value distributions, resulting from using $7$, $51$ and $100$ atoms, in the HopperBulletEnv-v0 environment}
    \label{fig:resolutionCompHopper}
\end{figure}

The reported results are averaged over 10 runs and 5 evaluation episodes per run and evaluation point. The reported average learning curves are much more similar in our setting than they are in the reported results from the discrete domain. In fact, as figures \ref{fig:paramCompAnt} and \ref{fig:resolutionCompHopper} show, there is not a clear advantage of any number of quantiles or their selection strategy over the others.

\begin{figure}
    \centering
    \includegraphics[width=\textwidth]{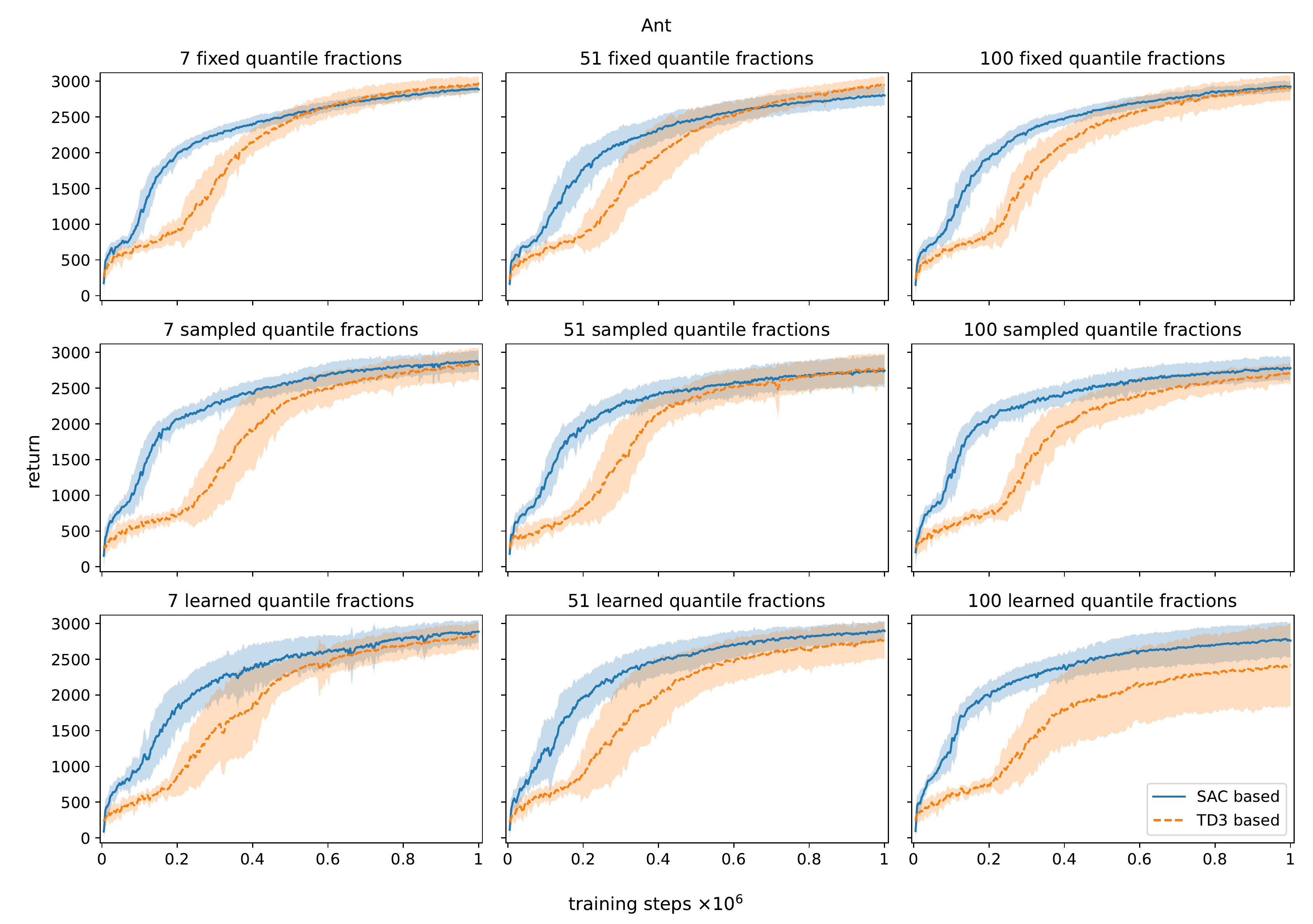}   
    \caption{Comparison of distributional algorithms based on either TD3 or SAC in the AntBulletEnv-v0 environment. SAC based agents improve more rapidly early on but final performance is similar.}
    \label{fig:base_alg_comp}
\end{figure}

Not only are the differences in learning speed and final performance not significant, but the best configuration also varies with the setting. As can be seen in figure \ref{fig:base_alg_comp} The SAC-based algorithms perform slightly better than their TD3 counterparts in most cases, supporting the choice of SAC over TD3 in \cite{duan_distributional_2021}. In the Ant environment specifically, they learn quicker early on.
While we do not dispute the general advantage of distributional over expected RL, our results suggest an invariance regarding the number and exact choice of quantiles. Note that all the environments considered here are deterministic except for the initial states of episodes. Prior work \cite{sun_interpreting_2022} supported the intuitive notion that learning value distributions instead of expected values could be helpful in stochastic or noisy (random or adversarial noise) environments. We therefore intend to extend this work to stochastic settings in the future.
This could yield further insight into the reasons for superior performance of distributional methods. The analysis in \cite{lyle_comparative_2019} shows that distributional RL is equal to expected RL in both a tabular as well as a linear function approximation setting and only beneficial with non-linear function approximation. \cite{sun_interpreting_2022} additionally show that the bounded gradient in a neural network-based agent helps make the agents more robust. A possible explanation for our results is that the benefits of distributional RL arise from the principle alone and require neither a large number or perfect distribution of quantiles, in deterministic environments. Further experiments should show whether the quality of approximation of value distributions has more influence in stochastic settings.

\subsection{Computation Time}
One factor that has not been considered above is the computation time. The IQN and FQF approaches bring with them a substantial increase in computation time. On the hardware used to run our experiments (NVIDIA GeForce RTX 3090 ($2\times$) and 3080 ($1\times$)) the two more complex methods used up to 5 times as much computation time as the basic quantile regression approach with fixed quantiles. This difference depends on the number of quantiles used. Other factors contribute to this, such as varying number of processes and therefore workload, or inefficiencies in our implementation of the distributional critics. More efficient implementations could probably reduce this discrepancy but the training of additional neural networks inevitably takes up significant computation time. \citeauthor{yang_fully_2019} report $20\%$ slower training for FQF compared to IQN, adding that FQF slows down further with increasing numbers of quantiles while IQN is not affected.

\section{Discussion and Future Work}\label{sec:Discussion}
We investigated three approaches to learning and representing value distributions in two widely used actor-critic algorithms. Our results indicate a qualitative invariance regarding the number and choice of quantile fractions predicted by the distributional critic. The significant improvements in performance from QR-DQN to IQN to FQF do not transfer straightforwardly to continuous action spaces using an actor-critic approach. Possible reasons for this include:
The learning dynamic between actor and critic: In the DQN based methods for the discrete action domain the policy is implicitly defined by the Q-function which directly profits from the distributional approach. In the continuous action actor-critic setting on the other hand the distributional critic does not dictate the policy and is only used to train the actor. Any improvement in the critic might not lead to an equivalent improvement in the actor.
The importance of implementation details and design choices of the discrete action space methods could be non-negligible. We made an effort to provide a fair comparison and used the same code for all critics where possible. This might have removed some sources of improvement.
It is also possible that the more complex methods are more sensitive to hyperparameter choice. The authors of FQF themselves state that it is not trivial to balance the learning rates of the different networks involved. We gave all algorithmic variants the same budget of hyperparameter sets to try out and also left some hyperparameters fixed across all of them. This might have negatively affected the FQF-based critic more than the fixed quantile critic.

Going forward, it would be interesting to see whether the presence of stochasticity in various elements of the environment changes these results. Despite the differing strategies for quantile fraction selection, the resulting sets of fractions are unlikely to degenerate (i.e. all fractions being very close to each other or in few clusters) and will therefore likely cover the probability space reasonably. In order to test how robust these quantile regression based methods really are to the choice of those fractions one could force agents to learn particularly bad sets of quantiles, i.e. by manipulating the sampling distribution in the IQN-based approach.

\bmhead{Acknowledgments}

This work is supported by the Ministry of Economics,
Innovation, Digitization and Energy of the State of
North Rhine-Westphalia and the European Union,
grants GE-2-2-023A (REXO) and IT-2-2-023 (VAFES)

\begin{appendices}
%\section{Implementation Details}\label{sec:implementationDetails}

\section{Hyperparameters}\label{sec:hyperparameters}
\renewcommand{\arraystretch}{1.5}
\begin{table}[h]\label{tab:commonHyperparams}
\begin{center}
\caption{Hyperparameters shared between all implemented algorithms}%
\begin{tabular}{|c|c|}
\hline
\textbf{Hyperparameter}     &       \textbf{Value}                \\
\hline \hline
critic network structure    & 2 hidden layers with 265 units each \\\hline
actor network structure     & 2 hidden layers with 265 units each \\\hline
activation function         & tanh \footnotemark[1]               \\\hline
batch size                  &   256                               \\\hline
$\kappa$ for quantile huber loss & $1.0$                          \\\hline
share feature extractor     & \multirow{2}*{True}                 \\
between actor and critic    &                                     \\\hline
number of critic networks   & 2                                   \\\hline
optimizer                   & Adam                                \\
\hline
\end{tabular}
\footnotetext[1]{except ReLu for distributional TD3 using fixed quantile fractions}
\end{center}
\end{table}

\begin{table}[h]\label{tab:commonHyperparamsCritics}
\begin{center}
\caption{Hyperparameters shared between all implemented algorithms where applicable}%
\begin{tabular}{|c|c|}
\hline
\textbf{Hyperparameter}     &       \textbf{Value}                \\
\hline \hline
number of cosines           & 64 \\\hline
total embedding dimension\footnotemark[1]   & 3136                \\\hline
entropy regularization      & \multirow{2}*{$0.05$}               \\
coefficient\footnotemark[2] &                                     \\\hline
FPN network structure\footnotemark[2] & single linear layer       \\\hline
FPN optimizer\footnotemark[2]         & RMSprop                   \\\hline
RMSprop alpha (TF: decay)\footnotemark[2] & $0.95$                \\\hline
RMSprop epsilon\footnotemark[2] & $0.00001$                       \\
\hline
\end{tabular}
\footnotetext[1]{applies to sampled and learned quantile fractions (IQN and FQF)}
\footnotetext[2]{applies to learned quantile fractions (IQN and FQF)}
\end{center}
\end{table}

\begin{table}[h]
\begin{center}
\caption{Hyperparameters of the Distributional TD3 Variants}\label{tab:hyperparamsTD3}%
\begin{tabular}{|l|ccc|}
\hline
  & fixed (QR-DQN) & sampled (IQN) & learned (FQF) \\
\hline
learning rate     & $4\cdot 10^{-4}$   & $2\cdot 10^{-4}$  & $2\cdot 10^{-4}$  \\\hline

FPN learning rate & - & - & $2\cdot 10^{-6}$  \\
\hline
\end{tabular}
\end{center}
\end{table}

\begin{table}[h]
\begin{center}
\caption{Hyperparameters of the Distributional SAC Variants}\label{tab:hyperparamsSAC}%
\begin{tabular}{|l|ccc|}
\hline
  & fixed (QR-DQN) & sampled (IQN) & learned (FQF) \\
\hline
learning rate     & $8\cdot 10^{-4}$   & $6\cdot 10^{-4}$  & $5\cdot 10^{-4}$  \\\hline

FPN learning rate & - & - & $5\cdot 10^{-6}$  \\
\hline
\end{tabular}
\end{center}
\end{table}




\end{appendices}

\clearpage
\bibliography{fgLit}% common bib file

\begin{thebibliography}{29}
\providecommand{\natexlab}[1]{#1}
\providecommand{\url}[1]{{#1}}
\providecommand{\urlprefix}{URL }
\providecommand{\doi}[1]{\url{https://doi.org/#1}}
\providecommand{\eprint}[2][]{\url{#2}}
 \bibcommenthead

\bibitem[{Akiba et~al(2019)Akiba, Sano, Yanase, Ohta, and
  Koyama}]{akiba_optuna_2019}
Akiba T, Sano S, Yanase T, et~al (2019) Optuna: {A} {Next}-generation
  {Hyperparameter} {Optimization} {Framework}. In: Proceedings of the 25rd
  {ACM} {SIGKDD} {International} {Conference} on {Knowledge} {Discovery} and
  {Data} {Mining}

\bibitem[{Barth-Maron et~al(2018)Barth-Maron, Hoffman, Budden, Dabney, Horgan,
  TB, Muldal, Heess, and Lillicrap}]{barth-maron_distributed_2018}
Barth-Maron G, Hoffman MW, Budden D, et~al (2018) Distributed {Distributional}
  {Deterministic} {Policy} {Gradients}. \doi{10.48550/arXiv.1804.08617},
  \urlprefix\url{http://arxiv.org/abs/1804.08617}, arXiv:1804.08617 [cs, stat]

\bibitem[{Bellemare et~al(2017)Bellemare, Dabney, and
  Munos}]{bellemare_distributional_2017}
Bellemare MG, Dabney W, Munos R (2017) A {Distributional} {Perspective} on
  {Reinforcement} {Learning}. \doi{10.48550/arXiv.1707.06887},
  \urlprefix\url{http://arxiv.org/abs/1707.06887}, arXiv:1707.06887 [cs, stat]

\bibitem[{Coumans and Bai(2016)}]{coumans_pybullet_2016}
Coumans E, Bai Y (2016) {PyBullet}, a {Python} module for physics simulation
  for games, robotics and machine learning. \urlprefix\url{http://pybullet.org}

\bibitem[{Dabney et~al(2018{\natexlab{a}})Dabney, Ostrovski, Silver, and
  Munos}]{dabney_implicit_2018}
Dabney W, Ostrovski G, Silver D, et~al (2018{\natexlab{a}}) Implicit {Quantile}
  {Networks} for {Distributional} {Reinforcement} {Learning}. arXiv:180606923
  [cs, stat] \urlprefix\url{http://arxiv.org/abs/1806.06923}, arXiv: 1806.06923

\bibitem[{Dabney et~al(2018{\natexlab{b}})Dabney, Rowland, Bellemare, and
  Munos}]{dabney_distributional_2018}
Dabney W, Rowland M, Bellemare M, et~al (2018{\natexlab{b}}) Distributional
  {Reinforcement} {Learning} {With} {Quantile} {Regression}. Proceedings of the
  AAAI Conference on Artificial Intelligence 32(1).
  \doi{10.1609/aaai.v32i1.11791},
  \urlprefix\url{https://ojs.aaai.org/index.php/AAAI/article/view/11791},
  number: 1

\bibitem[{Duan et~al(2021)Duan, Guan, Li, Ren, and
  Cheng}]{duan_distributional_2021}
Duan J, Guan Y, Li SE, et~al (2021) Distributional {Soft} {Actor}-{Critic}:
  {Off}-{Policy} {Reinforcement} {Learning} for {Addressing} {Value}
  {Estimation} {Errors}. IEEE Transactions on Neural Networks and Learning
  Systems pp 1--15. \doi{10.1109/TNNLS.2021.3082568},
  \urlprefix\url{http://arxiv.org/abs/2001.02811}, arXiv: 2001.02811

\bibitem[{Engel et~al(2005)Engel, Mannor, and Meir}]{engel_reinforcement_2005}
Engel Y, Mannor S, Meir R (2005) Reinforcement learning with {Gaussian}
  processes. In: Proceedings of the 22nd international conference on {Machine}
  learning. Association for Computing Machinery, New York, NY, USA, {ICML} '05,
  pp 201--208, \doi{10.1145/1102351.1102377},
  \urlprefix\url{https://doi.org/10.1145/1102351.1102377}

\bibitem[{Fujimoto et~al(2018)Fujimoto, Hoof, and
  Meger}]{fujimoto_addressing_2018}
Fujimoto S, Hoof H, Meger D (2018) Addressing {Function} {Approximation}
  {Error} in {Actor}-{Critic} {Methods}. In: Proceedings of the 35th
  {International} {Conference} on {Machine} {Learning}. PMLR, pp 1587--1596,
  \urlprefix\url{https://proceedings.mlr.press/v80/fujimoto18a.html}, iSSN:
  2640-3498

\bibitem[{Haarnoja et~al(2019)Haarnoja, Zhou, Hartikainen, Tucker, Ha, Tan,
  Kumar, Zhu, Gupta, Abbeel, and Levine}]{haarnoja_soft_2019}
Haarnoja T, Zhou A, Hartikainen K, et~al (2019) Soft {Actor}-{Critic}
  {Algorithms} and {Applications}. arXiv:181205905 [cs, stat]
  \urlprefix\url{http://arxiv.org/abs/1812.05905}, arXiv: 1812.05905

\bibitem[{Hassabis et~al(2017)Hassabis, Kumaran, Summerfield, and
  Botvinick}]{hassabis_neuroscience-inspired_2017}
Hassabis D, Kumaran D, Summerfield C, et~al (2017) Neuroscience-{Inspired}
  {Artificial} {Intelligence}. Neuron 95(2):245--258.
  \doi{10.1016/j.neuron.2017.06.011},
  \urlprefix\url{https://linkinghub.elsevier.com/retrieve/pii/S0896627317305093}

\bibitem[{Hessel et~al(2017)Hessel, Modayil, van Hasselt, Schaul, Ostrovski,
  Dabney, Horgan, Piot, Azar, and Silver}]{hessel_rainbow_2017}
Hessel M, Modayil J, van Hasselt H, et~al (2017) Rainbow: {Combining}
  {Improvements} in {Deep} {Reinforcement} {Learning}. arXiv:171002298 [cs]
  \urlprefix\url{http://arxiv.org/abs/1710.02298}, arXiv: 1710.02298

\bibitem[{Huber(1964)}]{kotz_robust_1964}
Huber PJ (1964) Robust {Estimation} of a {Location} {Parameter}. Annals of
  Mathematical Statistics pp 492--518. \doi{10.1007/978-1-4612-4380-9_35},
  \urlprefix\url{http://link.springer.com/10.1007/978-1-4612-4380-9_35}, book
  Title: Breakthroughs in Statistics ISBN: 9780387940397 9781461243809 Place:
  New York, NY Publisher: Springer New York

\bibitem[{Jaquette(1973)}]{jaquette_markov_1973}
Jaquette SC (1973) Markov {Decision} {Processes} with a {New} {Optimality}
  {Criterion}: {Discrete} {Time}. The Annals of Statistics 1(3):496--505.
  \doi{10.1214/aos/1176342415},
  \urlprefix\url{https://projecteuclid.org/journals/annals-of-statistics/volume-1/issue-3/Markov-Decision-Processes-with-a-New-Optimality-Criterion--Discrete/10.1214/aos/1176342415.full},
  publisher: Institute of Mathematical Statistics

\bibitem[{Koenker and Bassett~Jr(1978)}]{koenker_regression_1978}
Koenker R, Bassett~Jr G (1978) Regression quantiles. Econometrica: journal of
  the Econometric Society pp 33--50. Publisher: JSTOR

\bibitem[{Kuznetsov et~al(2020)Kuznetsov, Shvechikov, Grishin, and
  Vetrov}]{kuznetsov_controlling_2020}
Kuznetsov A, Shvechikov P, Grishin A, et~al (2020) Controlling {Overestimation}
  {Bias} with {Truncated} {Mixture} of {Continuous} {Distributional} {Quantile}
  {Critics}. In: Proceedings of the 37th {International} {Conference} on
  {Machine} {Learning}. PMLR, pp 5556--5566,
  \urlprefix\url{https://proceedings.mlr.press/v119/kuznetsov20a.html}, iSSN:
  2640-3498

\bibitem[{Li et~al(2018)Li, Bing, and Yang}]{li_distributional_2018}
Li S, Bing S, Yang S (2018) Distributional {Advantage} {Actor}-{Critic}.
  arXiv:180606914 [cs, stat] \urlprefix\url{http://arxiv.org/abs/1806.06914},
  arXiv: 1806.06914

\bibitem[{Lillicrap et~al(2019)Lillicrap, Hunt, Pritzel, Heess, Erez, Tassa,
  Silver, and Wierstra}]{lillicrap_continuous_2019}
Lillicrap TP, Hunt JJ, Pritzel A, et~al (2019) Continuous control with deep
  reinforcement learning. arXiv:150902971 [cs, stat]
  \urlprefix\url{http://arxiv.org/abs/1509.02971}, arXiv: 1509.02971

\bibitem[{Lyle et~al(2019)Lyle, Bellemare, and Castro}]{lyle_comparative_2019}
Lyle C, Bellemare MG, Castro PS (2019) A {Comparative} {Analysis} of {Expected}
  and {Distributional} {Reinforcement} {Learning}. Proceedings of the AAAI
  Conference on Artificial Intelligence 33:4504--4511.
  \doi{10.1609/aaai.v33i01.33014504},
  \urlprefix\url{https://aaai.org/ojs/index.php/AAAI/article/view/4365}

\bibitem[{Mnih et~al(2015)Mnih, Kavukcuoglu, Silver, Rusu, Veness, Bellemare,
  Graves, Riedmiller, Fidjeland, Ostrovski, Petersen, Beattie, Sadik,
  Antonoglou, King, Kumaran, Wierstra, Legg, and
  Hassabis}]{mnih_human-level_2015}
Mnih V, Kavukcuoglu K, Silver D, et~al (2015) Human-level control through deep
  reinforcement learning. Nature 518(7540):529--533. \doi{10.1038/nature14236},
  \urlprefix\url{http://www.nature.com/articles/nature14236}

\bibitem[{Morimura et~al(2012)Morimura, Sugiyama, Kashima, Hachiya, and
  Tanaka}]{morimura_parametric_2012}
Morimura T, Sugiyama M, Kashima H, et~al (2012) Parametric {Return} {Density}
  {Estimation} for {Reinforcement} {Learning}. \doi{10.48550/arXiv.1203.3497},
  \urlprefix\url{http://arxiv.org/abs/1203.3497}, arXiv:1203.3497 [cs, stat]

\bibitem[{Nam et~al(2021)Nam, Kim, and Park}]{nam_gmac_2021}
Nam DW, Kim Y, Park CY (2021) {GMAC}: {A} {Distributional} {Perspective} on
  {Actor}-{Critic} {Framework}. arXiv:210511366 [cs]
  \urlprefix\url{http://arxiv.org/abs/2105.11366}, arXiv: 2105.11366

\bibitem[{Nguyen-Tang et~al(2021)Nguyen-Tang, Gupta, and
  Venkatesh}]{nguyen-tang_distributional_2021}
Nguyen-Tang T, Gupta S, Venkatesh S (2021) Distributional {Reinforcement}
  {Learning} via {Moment} {Matching}. Proceedings of the AAAI Conference on
  Artificial Intelligence 35(10):9144--9152.
  \urlprefix\url{https://ojs.aaai.org/index.php/AAAI/article/view/17104},
  number: 10

\bibitem[{Raffin et~al(2021)Raffin, Hill, Gleave, Kanervisto, Ernestus, and
  Dormann}]{raffin_stable-baselines3_2021}
Raffin A, Hill A, Gleave A, et~al (2021) Stable-{Baselines3}: {Reliable}
  {Reinforcement} {Learning} {Implementations}. Journal of Machine Learning
  Research 22(268):1--8.
  \urlprefix\url{http://jmlr.org/papers/v22/20-1364.html}

\bibitem[{Silver et~al(2014)Silver, Lever, Heess, Degris, Wierstra, and
  Riedmiller}]{silver_deterministic_2014}
Silver D, Lever G, Heess N, et~al (2014) Deterministic {Policy} {Gradient}
  {Algorithms}. In: Proceedings of the 31st {International} {Conference} on
  {Machine} {Learning}. PMLR, pp 387--395,
  \urlprefix\url{https://proceedings.mlr.press/v32/silver14.html}, iSSN:
  1938-7228

\bibitem[{Sun et~al(2022)Sun, Zhao, Liu, Shi, Wang, Sadeghi, Yan, Jiang, and
  Kong}]{sun_interpreting_2022}
Sun K, Zhao Y, Liu Y, et~al (2022) Interpreting {Distributional}
  {Reinforcement} {Learning}: {Regularization} and {Optimization}
  {Perspectives}. \urlprefix\url{http://arxiv.org/abs/2110.03155},
  arXiv:2110.03155 [cs]

\bibitem[{Todorov et~al(2012)Todorov, Erez, and Tassa}]{todorov_mujoco_2012}
Todorov E, Erez T, Tassa Y (2012) {MuJoCo}: {A} physics engine for model-based
  control. In: 2012 {IEEE}/{RSJ} {International} {Conference} on {Intelligent}
  {Robots} and {Systems}. IEEE, pp 5026--5033, \doi{10.1109/IROS.2012.6386109}

\bibitem[{Witten(1977)}]{witten_adaptive_1977}
Witten IH (1977) An adaptive optimal controller for discrete-time {Markov}
  environments. Information and Control 34(4):286--295.
  \doi{10.1016/S0019-9958(77)90354-0},
  \urlprefix\url{https://linkinghub.elsevier.com/retrieve/pii/S0019995877903540}

\bibitem[{Yang et~al(2019)Yang, Zhao, Lin, Qin, Bian, and
  Liu}]{yang_fully_2019}
Yang D, Zhao L, Lin Z, et~al (2019) Fully {Parameterized} {Quantile} {Function}
  for {Distributional} {Reinforcement} {Learning}. In: Advances in {Neural}
  {Information} {Processing} {Systems}, vol~32. Curran Associates, Inc.,
  \urlprefix\url{https://proceedings.neurips.cc/paper/2019/hash/f471223d1a1614b58a7dc45c9d01df19-Abstract.html}

\end{thebibliography}


\end{document}